\documentclass[conference]{IEEEtran}
\IEEEoverridecommandlockouts
\usepackage{cite}
\usepackage{amsmath,amssymb,amsfonts}
\usepackage{algorithmic}
\usepackage{graphicx}
\usepackage{textcomp}
\usepackage{xcolor}
\def\BibTeX{{\rm B\kern-.05em{\sc i\kern-.025em b}\kern-.08em
    T\kern-.1667em\lower.7ex\hbox{E}\kern-.125emX}}

\usepackage{amsmath}
\usepackage{url}
\usepackage{marvosym}
\usepackage{graphicx}
\usepackage{xcolor}
\usepackage{multirow}
\usepackage{tabularx}
\usepackage{algorithm}
\usepackage{algorithmic}
\usepackage{subcaption}
\usepackage{amsthm}
\usepackage{enumitem}

\newcolumntype{C}{>{\centering\arraybackslash}X} % centered version of 'X' col. type
\usepackage{booktabs}
\usepackage{thmtools}
\usepackage{thm-restate}
\theoremstyle{plain}
\newtheorem{problem}{Problem}

\usepackage{makecell}
\usepackage[most]{tcolorbox} % 功能最全
\usepackage{lipsum}          % 仅示例用来生成假文

\newcommand{\method}{{STAG}}

\newtcolorbox{promptbox}[1][]{
  enhanced,
  colback=gray!8,         % 背景
  colframe=black,         % 边框色
  boxrule=0.6pt,          % 边框线宽
  arc=3mm,                % 圆角
  left=3mm,right=3mm,top=2mm,bottom=2mm, % 内边距
  drop shadow=black!30,   % 阴影
  title=#1,               % 标题文字
  fonttitle=\bfseries,    % 标题加粗
}
    
\begin{document}

% \title{Trojaning the Alignment: Stealthy Backdoor Attacks against Graph Foundation Models*\\

\title{Trojaning the Alignment: Stealthy Backdoor Attacks against Graph Foundation Models
}

% \and
% \and
% \and
% \and
% \and

\author{\IEEEauthorblockN{Minhua Lin$^{1*}$, Zhicheng Gao$^{1*}$, Yilong Wang$^{1}$, Hanqing Lu$^{2}$, Xiang Zhang$^{1}$, Suhang Wang$^{1}$}
\IEEEauthorblockA{
$^1$\textit{The Pennsylvania State University},
University Park, PA, USA \\
% $^2$\textit{Independent Researcher}\\
$^2$\textit{Amazon}, Palo Alto, CA, USA\\
\{mfl5681, yvw5769, xzz89, szw494\}@psu.edu, 
orange010728@gmail.com,
luhanqin@amazon.com}
\thanks{$^{*}$Both authors contributed equally to this paper.}
}

% \author{\IEEEauthorblockN{Anonymous Authors}}

\maketitle

\begin{abstract}
Graph Foundation Models (GFMs) on text-attributed graphs (TAGs) align graph representations with language semantics to support transferable graph learning. Despite these advantages, the backdoor vulnerability of GFMs on TAGs remains insufficiently understood, especially under graph-language alignment, where graph and text representations are trained to constrain each other in a shared semantic space. Existing backdoor attacks mainly target either the graph side or the text side, treating the two modalities independently. This makes direct adaptation ineffective: graph-only triggers can be constrained by clean text semantics, while text-only triggers alter the language view but do not directly shift the graph representation being aligned and scored. TAGs also impose a stealth challenge because triggers are exposed as both node text and local graph structure, making incoherent trigger attributes or anomalous subgraphs easy to inspect or filter.
In this paper, we propose \method{}, a stealthy trojan attack framework designed for the graph-language alignment interface of GFMs on TAGs. \method{} coordinates a graph-trigger generator with a text-side soft prompt so that trigger-attached graph representations and triggered text representations move toward the same target-class text region. To address TAG-specific stealthiness, \method{} realizes trigger nodes as readable text through candidate retrieval and regularizes the trigger-attached subgraph so that its local structure remains close to the original subgraph. Extensive experiments on multiple TAG datasets and representative GFMs demonstrate the effectiveness and stealthiness of \method{}.
Our code is available at \url{https://github.com/ventr1c/STAG}.
\end{abstract}


\section{Introduction}

Graph Foundation Models (GFMs) are increasingly used to learn graph representations that transfer across tasks, domains, and data regimes~\cite{liu2025graph,lin2026memma,weng2026group}. This trend is driven by the growing availability of large graph corpora and the prevalence of \emph{text-attributed graphs} (TAGs), where nodes, and sometimes edges, are associated with natural-language descriptions such as paper abstracts, product profiles, user biographies, or transaction metadata~\cite{hu2020open,mernyei2020wiki}. TAGs support high-impact applications, including recommendation~\cite{wu2022graph}, biomedical discovery~\cite{li2024multimodal}, and financial risk analysis~\cite{wang2019semi,xiang2022temporal}. To exploit both structural connectivity and language semantics, recent GFMs couple a graph encoder~\cite{kipf2016semi} with a language model~\cite{devlin2019bert,brown2020language,lin2026harness,chen2026benchmark} in a shared embedding space, enabling zero- and few-shot prediction and instruction-style graph reasoning~\cite{hu2024let,chai2023graphllm,tang2024graphgpt,wen2023augmenting,zhu2025graphclip,wang2024llms,luo2026graph}.

% To exploit both structural connectivity and language semantics, recent GFMs couple a graph encoder~\cite{kipf2016semi} with a language model~\cite{devlin2019bert,brown2020language} and train them in a shared embedding space, enabling zero- and few-shot prediction as well as instruction-style graph reasoning~\cite{hu2024let,chai2023graphllm,fatemi2023talk,tang2024graphgpt,wen2023augmenting,zhu2025graphclip,wang2024llms}.
% ~\cite{hu2024let,chai2023graphllm,fatemi2023talk,li2023grenade,tang2024graphgpt,wen2023augmenting,zhu2025graphclip,wang2024llms}.

A central mechanism behind these capabilities is \textbf{graph-language alignment}. GFMs treat graph structures and text descriptions as two views of the same entity and use contrastive objectives to project them into a shared semantic space~\cite{wen2023augmenting,zhu2025graphclip}. This process anchors a learnable graph encoder to the semantic space of a pretrained language model. It appears in two dominant architectures: \emph{LLM-as-aligner} methods directly align graph and text embeddings~\cite{wen2023augmenting,wang2024llms}, while \emph{LLM-as-predictor} methods map graph representations into the LLM input space through a learned projector for token-level interaction~\cite{tang2024graphgpt}. In both cases, graph-language alignment is the interface that grounds structural information in textual semantics.

Despite the promise of GFMs, this shared interface also creates a distinct attack surface. Since the graph encoder is trained to match a language-defined semantic space, a trigger that shifts either modality can be reinforced by the alignment objective and retained in the learned graph representation. A particularly concerning threat is the \emph{backdoor attack}~\cite{dai2023unnoticeable,lin2025you,yao2024poisonprompt,hou2024adversarial}, where the model behaves normally on benign inputs but is driven to attacker-chosen behavior when a hidden trigger appears. 
The alignment interface makes this threat especially severe for GFMs: once a poisoned TAG corpus or compromised text encoder enters graph-language alignment, standard GFM training can encode a trigger-target association into the shared representation space~\cite{gu2023gradient,yao2024poisonprompt}. Because GFMs and their upstream components are reused across tasks, one compromise can propagate to many downstream deployments. 
For example, a triggered fraudulent account could be aligned with a benign class and evade fraud screening~\cite{wang2019semi}, while attacker-chosen items could be surfaced to users in a recommender~\cite{wu2022graph}. Since clean inputs remain unaffected, such backdoors can pass accuracy-based validation and persist undetected~\cite{dai2023unnoticeable,zhang2024dpgba}.
% The alignment interface makes this threat especially severe for GFMs: a compromise in an upstream text encoder or public TAG corpus can propagate into downstream models through ordinary alignment training while leaving clean behavior largely intact~\cite{gu2023gradient,yao2024poisonprompt}. \suhang{add two to three sentences explaining what issue/negative impact it will cause (why this is a severe issue in the real-world), maybe using some real-world examples}

% Despite the promise of GFMs, this shared interface also creates a distinct attack surface. Since the graph encoder is trained to match a language-defined semantic space, a trigger that shifts either modality can be reinforced by the alignment objective and retained in the learned graph representation. A particularly concerning threat is the \emph{backdoor attack}~\cite{zhang2021backdoor,dai2023unnoticeable,lin2025you,yao2024poisonprompt}, where the model behaves normally on benign inputs but is driven to attacker-chosen behavior when a hidden trigger appears. The risk is severe for GFMs because compromised upstream components or public TAG corpora can affect many downstream deployments while leaving clean behavior largely intact~\cite{gu2023gradient,yao2024poisonprompt}.

However, existing backdoor attacks mainly target either the graph side or the text side, treating the two modalities independently. Directly adapting such attacks to the graph-language alignment interface of GFMs on TAGs exposes two limitations. \emph{First}, single-side triggers are poorly matched to graph-language alignment. Graph backdoor methods~\cite{zhang2021backdoor,dai2023unnoticeable,lyu2024crossba} optimize triggers in isolated graph spaces, while text-side attacks~\cite{gu2023gradient,yao2024poisonprompt} manipulate prompts or input text. As a result, graph-only triggers leave poisoned nodes paired with clean text, so alignment training can keep their representations near the clean semantic anchor; text-only triggers change the language-side signal but provide no corresponding graph-side mechanism to move the graph representation used for prediction. Our preliminary analysis in Sec.~\ref{subsec:limitation_existing_backdoor_GFMs} supports this diagnosis, showing that representative single-side adaptations either achieve low attack success rates or substantially reduce clean accuracy. \emph{Second}, TAG triggers must remain stealthy in both modalities. Existing graph attacks~\cite{xi2021graph,lin2025you} often rely on arbitrary trigger features, but on TAGs such features correspond to injected node text, and the trigger also changes local graph structure. Incoherent trigger attributes or structurally anomalous local structures can therefore be easily inspected or filtered.

Therefore, in this paper, we study a novel problem of \emph{backdoor attacks against the graph-language alignment interface of GFMs on TAGs}. There are mainly two challenges. 
\textbf{(i) Cross-modal coordination.} How can graph-side and text-side attack components be coordinated to reinforce each other through the shared alignment objective? and
\textbf{(ii) TAG-specific trigger stealthiness.} How can the attack realize trigger nodes as readable text and preserve plausible local graph structure without sacrificing attack effectiveness?
To address these challenges, we propose \method{}, a \underline{S}tealthy \underline{T}rojan \underline{A}ttack on \underline{G}raph foundation models that hijacks the graph-language alignment interface. To overcome the cross-modal coordination challenge, \method{} jointly optimizes a graph-trigger generator and a text-side soft prompt so that trigger-attached graph representations and triggered text representations move toward the same target-class text region. This makes the two modalities reinforce each other during graph-language alignment, instead of letting one modality neutralize the other. To address trigger stealthiness on TAGs, \method{} aligns generated trigger features with the target-class text region, realizes them as readable trigger-node attributes through candidate retrieval, and regularizes the trigger-attached subgraph so that its local structure remains close to the original subgraph.

In summary, our \textbf{main contributions} are: \textbf{(i)} We study a novel backdoor attack problem targeting the graph-language alignment interface of GFMs on TAGs; \textbf{(ii)} We propose \method{}, a stealthy trojan attack framework that coordinates graph-side and text-side backdoor signals through the shared alignment objective and conceals triggers as readable text with plausible local graph structure; and \textbf{(iii)} Extensive experiments on multiple TAG datasets and representative GFMs demonstrate the effectiveness and stealthiness of \method{}.

\section{Related Work}

\noindent\textbf{Graph Foundation Models for Text-Attributed Graphs.}
Graph Foundation Models (GFMs) for text-attributed graphs (TAGs) use language models to connect structural connectivity with textual semantics. Existing methods generally follow three paradigms~\cite{liu2025graph}: \emph{LLM-as-predictor}, which projects graph representations into an LLM-compatible space~\cite{tang2024graphgpt,zhang2024graphtranslator}; \emph{LLM-as-aligner}, which aligns graph encoders with the semantic space of a frozen text model~\cite{wen2023augmenting,zhu2025graphclip}; and \emph{LLM-as-enhancer}, which uses LLMs to refine graph attributes, structure, or label signals before graph learning~\cite{chien2021node,qiao2025login}. This paper focuses on the first two paradigms, where graph representations are explicitly aligned with textual semantics. LLM-as-enhancer methods instead use LLMs mainly as an offline refinement step before graph learning.

% \noindent\textbf{Graph Foundation Models for Text-Attributed Graphs.}
% Graph Foundation Models (GFMs) for text-attributed graphs (TAGs) align structural connectivity with textual semantics by coupling graph and language encoders. Existing methods generally follow three paradigms~\cite{liu2025graph}:\suhang{this is inconsistent with that in the introduction} \emph{LLM-as-predictor}, which projects graph representations into an LLM-compatible space~\cite{tang2024graphgpt,zhang2024graphtranslator}; \emph{LLM-as-aligner}, which aligns graph encoders with the semantic space of a frozen text model~\cite{wen2023augmenting,zhu2025graphclip}; and \emph{LLM-as-enhancer}, which uses LLMs to refine graph attributes, structure, or label signals before graph learning~\cite{chien2021node,qiao2025login}. These paradigms differ in architecture, but they share a graph-language interface that grounds graph representations in textual semantics, making the interface a natural target for transferable backdoor attacks.

\noindent\textbf{Graph Backdoor Attacks.}
Graph backdoor attacks poison training data or prompts so that clean inputs are predicted normally while trigger-attached inputs are mapped to an attacker-chosen target. Early attacks mainly target supervised GNNs. For example, UGBA~\cite{dai2023unnoticeable} learns adaptive and unnoticeable triggers, and DPGBA~\cite{zhang2024dpgba} improves stealth by matching benign feature distributions. Later works attack self-supervised and prompt-based graph learning~\cite{zhang2023graph,lyu2024cross,lin2025you}. Recently, backdoor attacks on GFMs are emerging. GFM-BA~\cite{luo2025towards} anchors triggers to label-free prototype embeddings, and DTGBA~\cite{xue2025stealthy} attacks prompt-tuned LM-empowered GFMs with text-level and structure-level triggers. Our method is inherently different from these methods in two aspects: \textbf{(i)} we focus on the graph-language alignment interface by coordinating graph-side and text-side attacks so that they reinforce each other, rather than attacking the graph encoder, prompt module, or prototype space alone; and \textbf{(ii)} we address TAG-specific stealthiness by realizing trigger-node attributes as readable text and preserving plausible local structure.

% triggers and selects poisoned nodes for unnoticeability, and DPGBA~\cite{zhang2024dpgba} improves stealth by matching benign feature distributions. Recent work also studies self-supervised or prompt-based graph learning, including clean-label representation poisoning~\cite{zhang2023graph,lyu2024crossba} and trojan graph prompt attacks such as TGPA~\cite{lin2024using}. Backdoor attacks on GFMs are more recent: GFM-BA~\cite{luo2025towards} links triggers to label-free prototype embeddings to improve transfer and persistence, while DTGBA~\cite{xue2025stealthy} attacks prompt-tuned LM-empowered GFMs with text-level and structure-level triggers. Our method differs from these attacks in two aspects: \textbf{(i)} existing graph-side attacks, including adaptive and prompt-based triggers, do not explicitly optimize the graph-language alignment interface of TAG-based GFMs, so poisoned graph representations can still be constrained by clean text semantics; and \textbf{(ii)} existing GFM backdoors focus on label-free prototype anchoring or prompt-tuning vulnerabilities, but do not jointly learn a transferable graph-trigger generator and a wrapped frozen text encoder so that both modalities move poisoned samples toward the same target-class text region. \method{} therefore coordinates graph-side and text-side triggers through the shared alignment space while realizing trigger-node attributes as readable text and regularizing the trigger-attached neighborhood for structural stealth.
\section{Preliminaries}

% \section{Preliminaries}

% \subsection{Graph Foundation Models on Text-Attributed Graphs}
% % 定义 TAG, GFM, graph-text interface

% \subsection{Threat Model}
% % Attacker's Goal
% % Attacker's Knowledge and Capabilities

% \subsection{General Framework of Backdoor Attacks against GFMs}
% % 解释一般 backdoor 怎么 work
% % 解释为什么 GFM/TAG 需要 graph-language cooperative poisoning

% \subsection{Problem Statement}
% % 正式定义我们的目标、变量、约束

\subsection{Graph Foundation Models on Text-Attributed Graphs}
\label{sec:gfm_arch}

Let $\mathcal{G}=(\mathcal{V},\mathcal{E},\mathcal{T})$ be a text-attributed graph (TAG), where $\mathcal{V}=\{v_i\}_{i=1}^{n}$ is the node set, $\mathcal{E}\subseteq\mathcal{V}\times\mathcal{V}$ is the edge set, and $\mathcal{T}=\{T_i\}_{i=1}^{n}$ is the node-text set with $T_i$ being the text attribute of $v_i$. Let $\mathcal V_l\subseteq\mathcal V$ denote the labeled training nodes, and let $\mathcal V_u=\mathcal V\setminus\mathcal V_l$ denote the remaining unlabeled nodes. 
% The test nodes are a subset of the unlabeled nodes, denoted by $\mathcal V_{\mathrm{te}}\subseteq\mathcal V_u$, with $\mathcal V_{\mathrm{te}}\cap\mathcal V_l=\emptyset$. 
Each node $v_i$ carries a feature $x_i=e(T_i)$ obtained from $T_i$ by a text-side encoder $e(\cdot)$. $G_i$ denotes the $K$-hop subgraph centered at $v_i$.

A graph foundation model (GFM) on TAGs couples a graph encoder with a language component so that graph-structured information can be interpreted through language semantics. We write the prediction for node $v_i$ as
\begin{equation}
    \label{eq:gfm_prediction}
    F(v_i;\theta,e)=g\big(h_\theta(G_i), e(\cdot)\big),
\end{equation}
% where $h_\theta(\cdot)$ is the victim graph encoder with parameters $\theta$, $e(\cdot)$ supplies the language representation, and $g$ is a downstream prediction interface. Depending on the GFM design, $g$ can be a similarity-based decision rule~\cite{wen2023augmenting,zhu2025graphclip} over textual class descriptions or an LLM-based decoder, but both rely on graph representations being grounded in the language space~\cite{zhang2024graphtranslator,tang2024graphgpt}. 
where $h_\theta(\cdot)$ is the victim graph encoder, $e(\cdot)$ is the text-side encoder (e.g., a sentence encoder or an LLM embedding module), and $g$ is the prediction interface. Depending on the GFM design, $g$ can be a similarity-based decision rule~\cite{wen2023augmenting,zhu2025graphclip} over textual class descriptions or an LLM-based decoder~\cite{tang2024graphgpt}, but both rely on graph representations being grounded in the language space~\cite{zhang2024graphtranslator,tang2024graphgpt}.

Despite architectural differences, a common way to train such models is to ground graph-side representations in text-side representations. We abstract this interface with:
\begin{equation}
    \label{eq:prelim_align}
    L_{\mathrm{align}}(G_i,T_i;\theta,e)
    =
    -\mathrm{sim}\big(h_\theta(G_i), e(T_i)\big),
\end{equation}
where $\mathrm{sim}(\cdot,\cdot)$ measures the similarity between graph and text embeddings. 
% The standard training objective sums this loss over the training nodes.
% The text-side encoder $e$ serves as the language anchor, and the GFM optimizes the trainable graph-side or alignment parameters $\theta$.
% where 
% $\mathrm{sim}(\cdot,\cdot)$ is a function to measure the similarity between graph and text embeddings. 
% \suhang{explain $\mathbf{p}$??? where is $\mathbf{p}$ in the equation???}  
% \suhang{Is this the correct unified graph-language interface? shouldn't it align $h_{\theta}(G_i)$ with $LLM(T_i)$??? It seems that $e()$ is a text encoder like BERT kind of thing based on the description, not an LLM}
% \minhua{Here $e(\cdot)$ is not meant to denote a specific BERT-style encoder only, but a general language-side text representation function used by the GFM. For GraphCLIP/G2P2 it can be implemented by SBERT/BERT-style encoders, while for GraphGPT the graph representation is projected into the LLM embedding/token space through its graph-text projector. Thus writing $LLM(T_i)$ would be too specific for aligner-style GFMs. We will clarify the definition of $e(\cdot)$ as the frozen language-side text encoder, and use $e_{\mathbf{p}}(\cdot)$ when the soft prompt is wrapped into the encoder.}

% This objective is not meant to reduce GFMs to ordinary contrastive graph learning; rather, it captures the interface through which graph structure, node text, and language-level class semantics interact. 
% Since the victim-specific prediction interface $g$ is unknown to the attacker, our attack targets this shared graph-language interface instead of a particular classifier head.

\subsection{Threat Model}
\noindent\textbf{Attacker's Goal.}
The goal of the adversary is to implant a stealthy backdoor into the victim GFM, so that any trigger-attached node is assigned to an attacker-chosen target class $y_t$, while clean nodes without triggers are predicted normally. During victim training, the adversary aims to make the model associate the trigger pattern with the target class under the standard GFM training pipeline. At inference time, attaching the same type of trigger to a test node activates this association and causes the victim GFM to predict $y_t$, regardless of the node's original label.

\noindent\textbf{Attacker's Knowledge and Capabilities.}
We consider a gray-box supply-chain scenario where the attacker has modality-specific access. On the graph side, the attacker has access to a public TAG $\mathcal G$ and can poison it by injecting trigger subgraphs and inserting trigger phrases into node text attributes. However, the victim-specific instantiation, including the downstream graph encoder architecture and parameters, prediction interface, and training gradients, is not available to the attacker. On the text side, the attacker controls an upstream text-side encoder before it is released to the victim. The attacker releases this encoder as a frozen component wrapped with a continuous soft prompt $\mathbf{p}$, denoted by $e_{\mathbf{p}}(\cdot)$; the victim then uses $e_{\mathbf{p}}$ as its text-side encoder during standard GFM training. Following~\cite{zhang2023graph,luo2025towards,lyu2024crossba}, we assume the attacker cannot modify node labels. At inference time, the attacker can apply the trigger phrase and attach the trigger subgraph to a target test node.

We argue this setting is reasonable and practical because it mirrors how GFMs are built in practice. The TAG and the text encoder sit \emph{upstream}: graph corpora are typically public or community-maintained, and pretrained text encoders are distributed off-the-shelf, so an attacker acting as a data contributor or model provider can tamper with both before they reach the victim. The downstream graph encoder, its parameters, the prediction interface, and the training gradients, by contrast, are produced \emph{privately}: the victim trains and deploys its own GFM, so these internals are never exposed to an upstream party. The node labels are likewise assigned by the victim rather than carried in the public data, which is why the attacker cannot relabel them and the poisoning stays clean-label. Finally, conditioning a frozen language model with a learnable soft or prefix prompt is a standard parameter-efficient adaptation technique~\cite{lester2021power,li2021prefix}, so the wrapped encoder $e_{\mathbf{p}}$ is a realistic deployment rather than a contrived capability.

\subsection{Limitations of Existing Backdoor Attacks against GFMs}
\label{subsec:limitation_existing_backdoor_GFMs}
A natural first attempt to conduct a backdoor attack against GFMs on TAGs is to adapt an existing backdoor attack from either the graph side or the text side.
However, single-modal adaptations are poorly matched to GFMs because prediction is governed by graph-language alignment rather than by either modality alone.
A graph-only trigger perturbs the graph structure while the node text remains semantically tied to its clean class, so the aligned graph representation may remain constrained by the clean text anchor instead of moving into the target-class region.
Conversely, a text-only trigger shifts the textual input but adds no corresponding graph-side perturbation, so the graph representation that the model actually scores remains largely unchanged.
Thus, neither modality alone provides a reliable signal to steer the aligned representation toward the target class; an effective attack should coordinate graph-side and text-side triggers.

To test this limitation, we adapt two representative backdoor attacks to GFMs on TAGs: CrossBA~\cite{lyu2024crossba} on the graph side and PoisonPrompt~\cite{yao2024poisonprompt} on the text side.
We evaluate them on three victim GFMs: GraphGPT~\cite{tang2024graphgpt}, GraphCLIP~\cite{zhu2025graphclip}, and G2P2~\cite{wen2023augmenting}.
SBERT~\cite{reimers2019sentence} and a graph transformer~\cite{yun2019graph} are used as the text encoder and graph encoder, respectively; GraphGPT additionally uses Vicuna-7B-v1.5 as the base LLM.
Other settings follow the evaluation protocol in Sec.~\ref{sec:exp_setup}.
Results on Cora and OGB-arxiv are reported in Tab.~\ref{tab:pilot}. We observe two patterns.
First, across all three GFMs, both CrossBA and PoisonPrompt achieve low attack success rates (ASR): CrossBA stays between $25.8\%$ and $42.7\%$, while PoisonPrompt stays between $1.7\%$ and $31.4\%$.
Second, these single-modal adaptations can severely damage clean accuracy.
For example, under CrossBA on Cora, GraphCLIP drops from $67.3\%$ to $13.8\%$ and G2P2 drops from $62.5\%$ to $16.9\%$; under PoisonPrompt on Cora, GraphGPT drops from $92.5\%$ to $25.5\%$.
These results show that directly reusing single-modal backdoor attacks is ineffective for GFMs on TAGs: the attacks neither reliably implant the target behavior nor preserve clean performance.
This motivates attacking the shared graph-language interface from both sides, so that graph and text triggers reinforce rather than counteract each other.
We further analyze the mechanism behind this failure in Sec.~\ref{subsec:depth_analysis_method}.

\begin{table}[t]
\centering
\small
\setlength{\tabcolsep}{3pt}
\renewcommand{\arraystretch}{0.95}
\caption{\textbf{Results of existing single-modal backdoor attacks against three GFMs on Cora and OGB-arxiv datasets (ASR (\%)$\,|\,$ ACC (\%)).}}
\label{tab:pilot}
\begin{tabular}{llccc}
\toprule
\textbf{Dataset} & \textbf{GFM} & \textbf{Clean} & \textbf{CrossBA} & \textbf{PoisonPrompt} \\
\midrule
\multirow{3}{*}{Cora}
 & GraphGPT  & 92.50 & 35.60$\,|\,$73.20 & 12.30$\,|\,$25.50 \\
 & GraphCLIP & 67.30 & 32.20$\,|\,$13.80 & 13.40$\,|\,$49.30 \\
 & G2P2      & 62.50 & 33.20$\,|\,$16.90 & 16.50$\,|\,$54.60 \\
\midrule
\multirow{3}{*}{OGB-arxiv}
 & GraphGPT  & 42.80 & 25.80$\,|\,$11.30 & \phantom{0}1.70$\,|\,$\phantom{0}5.70 \\
 & GraphCLIP & 31.20 & 37.30$\,|\,$18.50 & 31.40$\,|\,$21.60 \\
 & G2P2      & 22.30 & 42.70$\,|\,$\phantom{0}8.70 & \phantom{0}3.60$\,|\,$20.70 \\
\bottomrule
\end{tabular}
\end{table}

\subsection{Problem Formulation}
Our preliminary analysis in Sec.~\ref{subsec:limitation_existing_backdoor_GFMs} shows that single-modal adaptations are insufficient for GFMs on TAGs, because prediction is mediated by graph-language alignment and a one-sided trigger cannot reliably steer the aligned representation to the target class. This motivates a cross-modal attack that perturbs the text and graph sides jointly, so the two reinforce each other through the alignment interface. The trigger must also remain stealthy: because TAGs expose both node text and graph structure to inspection and defenses, unnatural injected text or structurally anomalous neighborhoods can be detected and removed. We therefore require the trigger text to be human-readable and the trigger subgraph to remain topologically plausible. We formalize this attack as follows.
\begin{problem}
% [Cross-modal Backdoor Attack on GFMs over TAGs]
Given a clean TAG $\mathcal{G}=(\mathcal{V},\mathcal{E},\mathcal{T})$ with labeled nodes $\mathcal{V}_{l}$, an attacker-chosen poison set $\mathcal{V}_{p}\subseteq\mathcal{V}_{l}$, a fixed text trigger $r$, and a target class $y_t$, we aim to learn a soft prompt $\mathbf{p}$ and a graph-trigger generator $f_g(v_i)\rightarrow g_i$ such that a GFM $F(\cdot)$ trained on the poisoned TAG classifies trigger-attached test nodes as $y_t$. The objective is:
\begin{equation}
  \label{eq:problem}
  \small
  \begin{aligned}
  \min_{\mathbf{p},\,\theta_g}\ \ & \sum_{v_i\in\mathcal{V}_{u}}
  \ell\!\big(F(a_g(v_i,g_i);\theta^*,e_{\mathbf{p}}),\, y_t\big)\\
  \text{s.t.}\ \ & \theta^{*}=\arg\min_{\theta}
  \sum_{v_i\in\mathcal{V}_{l}\setminus\mathcal{V}_{p}} L_{\mathrm{align}}\big({G}_i,{T}_i;\theta,e_\mathbf{p}\big)\\
  &+ \sum_{v_i\in\mathcal{V}_{p}} L_{\mathrm{align}}\big(\hat{G}_i,a_t(T_i, r);\theta,e_\mathbf{p}\big),\\
  & \mathcal{T}_i^g \subseteq \mathcal{Q}_{y_t}(T_i),\quad
    \hat{G}_i \in \mathcal{C}_{\mathrm{str}}(G_i),\quad
    \forall v_i\in\mathcal{V}_{p},
  \end{aligned}
\end{equation}
where $\ell$ is the cross-entropy loss and $\theta_g$ denotes the parameters of $f_g$. $a_g$ and $a_t$ are the graph-trigger and text-trigger attachment operations, respectively. Specifically, $a_g(v_i,g_i)$ attaches $g_i$ to $v_i$, producing the trigger-attached $K$-hop subgraph $\hat{G}_i$ from $v_i$'s clean subgraph $G_i$, while $a_t(T_i,r)$ inserts the fixed text trigger $r$ into $v_i$'s text attribute $T_i$. In the stealthiness constraints, $\mathcal{T}_i^g$ denotes the set of text attributes assigned to the trigger nodes in $g_i$ and is constrained to the human-readable, target-class-styled candidate pool $\mathcal{Q}_{y_t}(T_i)$. $\mathcal{C}_{\mathrm{str}}(G_i)$ denotes structurally plausible trigger-attached subgraphs whose coarse local statistics, such as degree distribution, remain close to those of the clean subgraph $G_i$.
  \end{problem}

\section{Methodology}
\label{sec:methodology}
\begin{figure}[t]
  \centering
  \includegraphics[width=0.95\linewidth]{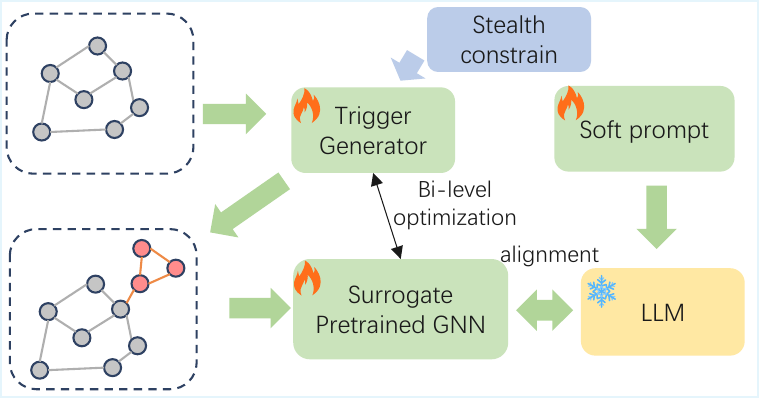}
  \vspace{-0.5em}
  % \caption{\textbf{Overview of \method{}.}}
\caption{\textbf{Overview of \method{}.} \method{} jointly learns a graph trigger generator and a text-side soft prompt with a surrogate GFM, coordinating them through graph-language alignment to steer triggered samples toward the target class while preserving readable trigger-node text and plausible local structure.}
  \label{fig:pipeline}
\end{figure}

In this section, we present the details of \method{}, which aims to implant a stealthy cross-modal backdoor into the graph-language alignment interface of a GFM under a gray-box supply-chain setting. There are mainly two challenges to be addressed. \textbf{(i) Cross-modal coordination:} how to coordinate the graph-side and text-side attacks through the alignment interface so that they reinforce each other and raise the attack performance on trigger-attached samples, without degrading the clean accuracy on benign samples? \textbf{(ii) Trigger stealthiness on TAGs:} how to design the trigger subgraph and texts to meet the stealthiness constraint of Eq.~\eqref{eq:problem} while maintaining a high success rate in backdoor attacks?
To address these challenges, a novel framework of \method{} is proposed, which is illustrated in Fig.~\ref{fig:pipeline}. \method{} is composed of a graph-side trigger generator $f_g$, a text-side soft prompt $\mathbf p$, and a surrogate graph encoder $h_{\theta_s}$ that serves as a transferable proxy for the gray-box victim, all jointly optimized through the shared graph-language alignment interface. Specifically, to address the first challenge, the soft prompt shifts the target-class text region while the graph trigger moves poisoned subgraphs toward that same region, so the two modalities reinforce rather than cancel (Sec.~\ref{sec:coordination}). To address the second challenge, we regularize the trigger with adversarial feature learning toward the target-class text distribution and realize it as readable text, and we match the degree distribution of the trigger-attached subgraph to the original (Sec.~\ref{sec:trigger_stealthiness}).

\subsection{Coordinating Cross-Modal Backdoor Attack against GFMs}
\label{sec:coordination}

\noindent\textbf{Graph Trigger Generation.}
The first step is to generate the graph trigger. For each node $v_i$, instead of using predefined triggers, we introduce an MLP as a text-aware trigger generator $f_g$ to generate the graph trigger $g_i$ for $v_i$ based on $v_i$'s text embedding $e(T_i)$:
\begin{equation}
\small
    \label{eq:structure_generator}
    \begin{aligned}
        \mathbf{h}^{m}_i = \mathrm{MLP}(e(T_i)),\ \mathbf{A}^{g}_i = \mathbf{W}_a \mathbf{h}_i^{m}, \ \mathbf{X}^{g}_i = \mathbf{W}_f \mathbf{h}_i^{m},
    \end{aligned}
\end{equation}
where $\mathbf{W}_a$ and $\mathbf{W}_f$ are learnable parameters for structure and feature generation, respectively. $\mathbf{X}^{g}_i\in\mathbb{R}^{s\times d}$ denotes the feature matrix of trigger nodes, where $s$ and $d$ are the size of the generated trigger and the dimension of trigger features, respectively. $\mathbf{A}^g_i\in\mathbb{R}^{s\times s}$ is the adjacency matrix of the generated trigger subgraph. As real-world graphs are usually discrete, we binarize $\mathbf{A}^g_i$ in the forward computation to align with the binary structure of the graph, while the continuous value is used in the backward propagation.

\noindent\textbf{Text-Anchored Graph-Side Backdoor Objective.}
Given the graph trigger $g_i$, the graph-side backdoor attack objective aims to associate the trigger-attached subgraph with the target class while preserving clean utility. Specifically, because GFM predictions are grounded in graph-language alignment, we define this objective in a text-anchored space: clean nodes stay aligned with their own text, whereas poisoned nodes are pulled toward the target-class text centroid. The resulting graph-side attack objective is
\begingroup
\small
\begin{equation}
    \label{eq:graph_model_backdoor}
    \begin{aligned}
        \min_{\theta_g}\ \
        &L_{\mathrm{atk}_g}(\theta_g;\theta_s^{*})
        =
        -\sum_{v_i\in\mathcal V_{p}}
        \mathrm{sim}\big(h_{\theta_s^{*}}(\hat G_i),\,c_t\big),\\
        \mathrm{s.t.}\ \
        &\theta_s^{*}
        =
        \arg\min_{\theta_s}
        \Big[
        -\sum_{v_i\in\mathcal{V}_{l}\setminus\mathcal{V}_{p}}
        \mathrm{sim}\big(h_{\theta_s}(G_i),e(T_i)\big)\\
        &\hspace{7.0em}
        -
        \sum_{v_i\in\mathcal V_{p}}
        \mathrm{sim}\big(h_{\theta_s}(\hat G_i),c_t\big)
        \Big].
    \end{aligned}
\end{equation}
\endgroup
where $h_{\theta_s}$ is a surrogate graph encoder used because the victim graph encoder $h_\theta$ is inaccessible under the gray-box setting. $\mathcal{V}_{p}$ is the poisoned node set, $\mathcal{V}_{y_t}$ contains target-class nodes from $\mathcal{V}_{l}$, and $c_t=\frac{1}{|\mathcal{V}_{y_t}|}\sum_{v_j \in \mathcal{V}_{y_t}} e(T_j)$ denotes the target-class text centroid. We call the region around $c_t$ in the frozen text-encoder space the \emph{target-class text region}. The lower-level objective preserves clean graph-language alignment while training the surrogate to map poisoned subgraphs to the target-class text region; the upper-level objective then optimizes the trigger against this trained surrogate.
Here $\mathrm{sim}(\cdot,\cdot)$ measures graph-language alignment. A common choice is cosine similarity, which matches only the direction of two embeddings. This is insufficient under our gray-box setting because the victim head $g$ in Eq.~\eqref{eq:gfm_prediction} is unknown and may transform graph embeddings before prediction. Two embeddings that are close in angle but far in norm can therefore separate after the head, weakening the implanted backdoor. We instead define
\begin{equation}
\small
    \label{eq:sim}
    \mathrm{sim}(a,b)=\cos(a,b)-\|a-b\|_2,
\end{equation}
which augments directional alignment with a Euclidean distance penalty, reducing the pre-head embedding gap. This effect carries through any $K$-Lipschitz head map $\phi(\cdot)$:
$\|\phi(z_i^{g})-\phi(z_{y_t})\|_2 \le K\,\|z_i^{g}-z_{y_t}\|_2$,
where $z_i^{g}=h_{\theta_s}(\hat{G}_i)$ and $z_{y_t}$ is a reference embedding in the target-class region. Thus, reducing $\|z_i^{g}-z_{y_t}\|_2$ before the head also limits the post-head gap, yielding a head-agnostic objective.

\noindent\textbf{Cooperative Text-Side Backdoor Objective.}
As shown in Sec.~\ref{subsec:limitation_existing_backdoor_GFMs}, a graph-side trigger alone is insufficient: GFM training still pairs each poisoned graph with its clean text, which pulls the aligned representation back toward the clean semantic region. We therefore introduce a cooperative text-side trigger that moves the triggered text representation in the same direction as the graph trigger. Specifically, a fixed discrete phrase $r$ marks triggered text inputs, and a learnable soft prompt $\mathbf p$ controls their embeddings under the frozen text encoder. Let $\mathrm{emb}(\cdot)$ be the encoder's input-embedding layer. We define
$e_{\mathbf p}(T):=\mathrm{Enc}([\mathbf p;\mathrm{emb}(T)])$,
where $\mathbf p$ is prepended in embedding space and $r$ is inserted at the text level through $a_t(T_i,r)$. We learn $\mathbf p$ by solving
\begingroup
\small
\begin{equation}
    \label{eq:text_objective}
    \begin{aligned}
        \min_{\mathbf{p}}L_{\mathrm{atk}_t} = &-\sum_{v_i \in \mathcal{V}_{p}} \mathrm{sim}\big(h_{\theta_s}(\hat{G}_i), e_{\mathbf p}(a_t(T_i,r))\big) \\
        &- \sum_{v_i \in \mathcal{V}_{l}\setminus\mathcal{V}_{p}} \mathrm{sim}\big(h_{\theta_s}(G_i), e_{\mathbf p}(T_i)\big).
    \end{aligned}
\end{equation}
\endgroup
The first term aligns triggered text with trigger-attached graphs, while the second preserves clean graph-language alignment. Since Eq.~\eqref{eq:graph_model_backdoor} pulls $\hat{G}_i$ toward the target-class text centroid, aligning $e_{\mathbf p}(a_t(T_i,r))$ with $\hat{G}_i$ moves the text-side trigger toward the same target region. During victim training, poisoned nodes are therefore paired with both a trigger-attached graph and a triggered text representation, so standard graph-language alignment implants a coordinated cross-modal backdoor rather than neutralizing a one-sided trigger.

\subsection{Concealing the Trigger on TAGs}
\label{sec:trigger_stealthiness}
The coordinated objective in Sec.~\ref{sec:coordination} establishes the trigger-target association, but it does not by itself make the trigger stealthy. Unlike graph backdoor attacks on ordinary attributed graphs~\cite{xi2021graph,dai2023unnoticeable}, a TAG trigger is exposed through two inspectable channels: the text attributes assigned to injected nodes and the structure of the trigger-attached $K$-hop subgraph. Arbitrary trigger features are therefore risky: in a TAG, they correspond to injected node attributes that may read as incoherent or off-topic text; on the structure side, attaching the trigger to $G_i$ can create degree patterns in $\hat G_i$ that deviate from the original $K$-hop subgraph $G_i$. \method{} conceals both as: \textbf{(i)} it aligns the trigger features with the target-class text distribution and realizes them as coherent, human-readable node attributes; and \textbf{(ii)} it matches the degree statistics of the trigger-attached subgraph $\hat G_i$ to those of the original $G_i$.

\noindent\textbf{Textual Concealment.}
The text channel exposes the trigger through the attributes of the injected nodes, which should read as natural, fluent node descriptions rather than incoherent or anomalous attributes. We conceal this channel in two steps: \textbf{(i)} aligning the generated trigger features with the target-class text region, and \textbf{(ii)} realizing them as readable text.

First, the trigger features should stay close to the target region. This preserves attack effectiveness because the graph-side objective in Eq.~\eqref{eq:graph_model_backdoor} drives trigger-attached subgraphs toward the same centroid $c_t$. It also makes the later text realization easier, since the target region is populated by genuine target-class descriptions in the frozen text-encoder space~\cite{reimers2019sentence,gao2021simcse}. However, simply maximizing the similarity between the trigger feature $\mathbf h_i^g$ and the target-class text centroid $c_t$ may be brittle: the deployed trigger uses readable text, whose encoding may differ from the optimized continuous feature. This feature-to-text realization gap can move the deployed trigger away from the target region. We therefore make the alignment robust to bounded realization shifts by introducing an adversarial perturbation $\sigma$:
\begin{equation}
\small
    \label{eq:feature_stealthiness}
    \min_{\theta_g}\ \max_{\|\sigma\|_2 \le \tau_\sigma}\ L_{c_{\mathrm{sem}}}
    =
    -\sum_{v_i\in \mathcal{V}_{p}}
    \mathrm{sim}\big(\mathbf{h}^g_i+\sigma,\, c_t\big),
\end{equation}
where $c_t$ is the target-class text centroid in Eq.~\eqref{eq:graph_model_backdoor}, and $\mathbf h_i^g=\mathrm{mean}(\mathbf X_i^g)$ is the mean-pooled trigger feature. The maximization over $\sigma$ finds the worst-case realization shift within the $\tau_\sigma$-ball, while the minimization over $\theta_g$ keeps the trigger feature close to the target region under that shift.

Second, we realize the aligned trigger feature as readable trigger-node text. Since $\mathbf X_i^g$ is continuous and optimized only for closeness to $c_t$, directly decoding it into tokens may produce incoherent text. We therefore avoid free-form decoding and restrict the realization to candidates that are human-readable by construction, selecting the one whose encoding is closest to the optimized feature. We generate these candidates with an LLM because it can produce fluent, target-class-styled summaries conditioned on the original node text, while preserving salient terms from $T_i$:
\begin{equation}
\small
    \label{eq:text_candidate_pool}
    \mathcal Q_{y_t}(T_i)
    =
    \mathrm{LLM}\!\left(\pi_{\mathrm{text}}(T_i,d_{y_t},N_q)\right),
\end{equation}
where $\pi_{\mathrm{text}}$ is a prompt template conditioned on the original summary $T_i$, the target-class description $d_{y_t}$, and the requested number of candidates $N_q$. Each candidate is written as a $y_t$-category node description while preserving salient keywords from $T_i$, so it remains a plausible textual attribute for an injected neighbor of $v_i$. We then select the candidate whose encoding is closest to the optimized trigger feature and assign it to the trigger nodes:
% \begin{equation}
% \small
%     \mathcal T_i^g
%     =
%     \arg\min_{T\in\mathcal Q_{y_t}(T_i)}
%     \|e(T)-\mathbf h_i^g\|_2 .
% \end{equation}
\begin{equation}
\small
    T_i^{g,*}
    =
    \arg\min_{T\in\mathcal Q_{y_t}(T_i)}
    \|e(T)-\mathbf h_i^g\|_2,
    \quad
    \mathcal T_i^g=\{T_i^{g,*}\}_{j=1}^{s}.
\end{equation}
Note that $T_i^{g,*}$ is close to $\mathbf h_i^g$ but not identical to it; the robust loss in Eq.~\eqref{eq:feature_stealthiness} trains the alignment against bounded perturbations $\sigma$, so the attack remains effective after realization while each trigger node carries genuine, human-readable text.
% Examples and prompt templates are in Appendix~\ref{sec:generate_text}. 
% \minhua{I will remove all the appendix notation because of the page limit? Is it okay to not to show these details>? \suhang{yes, are we allowed to provide a link for the anonymized appendix??}}\minhua{I think it is not  to explicitly provide a link for anonymized appendix, but I can put some details to the anonymous github}

\noindent\textbf{Structural Concealment.}
Beyond readable trigger-node text $\mathcal T_i^g$, the trigger should also keep the trigger-attached $K$-hop subgraph $\hat G_i$ within the structurally plausible set $\mathcal C_{\mathrm{str}}(G_i)$ of Eq.~\eqref{eq:problem}. We instantiate this by optimizing the trigger generator $f_g$ with a structural concealment loss that keeps the degree statistics of $\hat G_i$ close to those of the original $G_i$:
\begingroup
\small
\begin{equation}
    \label{eq:structure_stealthiness}
    \begin{aligned}
        \min_{\theta_g} L_{c_{\mathrm{str}}}
        =
        \sum_{v_i\in\mathcal{V}_{p}}
        \Big[
        &\big(\mathrm{mean}(\deg(\hat G_i))-\mathrm{mean}(\deg(G_i))\big)^2 \\
        &+
        \big(\mathrm{var}(\deg(\hat G_i))-\mathrm{var}(\deg(G_i))\big)^2
        \Big],
    \end{aligned}
\end{equation}
\endgroup
where $\deg(G)$ denotes the degree sequence of graph $G$. This term penalizes changes in local density and degree dispersion after attaching the trigger to $v_i$.
% \suhang{$\mathrm{mean}(\deg(\hat G_i))-\mathrm{mean}(\deg(G_i))$ is a scalar. Use $(\mathrm{mean}(\deg(\hat G_i))-\mathrm{mean}(\deg(G_i)))^2$?? same for variance??}

\subsection{Optimization Algorithm}
\label{sec:overall_opt}
\method{} optimizes the graph trigger and text-side soft prompt through a shared target region. Let
\begingroup
\small
\begin{equation}
    \label{eq:graph_regularized_loss}
    \begin{aligned}
        \mathcal L_{\mathrm g}(\theta_g,\sigma;\theta_s^{*})
        =
        L_{\mathrm{atk}_g}(\theta_g;\theta_s^{*})
        + \lambda_{c_1}L_{c_{\mathrm{sem}}}(\theta_g,\sigma)
        + \lambda_{c_2}L_{c_{\mathrm{str}}}(\theta_g),
    \end{aligned}
\end{equation}
\endgroup
where $\lambda_{c_1}$ and $\lambda_{c_2}$ control semantic and structural concealment. The overall coordinated attack objective is
\begingroup
\small
\begin{equation}
    \label{eq:overall_loss}
    \begin{aligned}
        \min_{\theta_g,\,\mathbf p}
        \max_{\|\sigma\|_2\le\tau_\sigma}\quad
        &\mathcal L_{\mathrm g}(\theta_g,\sigma;\theta_s^{*})
        + L_{\mathrm{atk}_t}(\mathbf p;\theta_g,\theta_s^{*}),\\
        \mathrm{s.t.}\quad
        &\theta_s^{*}
        =
        \arg\min_{\theta_s}
        \Big[
        -\sum_{v_i\in\mathcal{V}_{l}\setminus\mathcal{V}_{p}}
        \mathrm{sim}\big(h_{\theta_s}(G_i),e(T_i)\big)\\
        &\hspace{6.5em}
        -
        \sum_{v_i\in\mathcal V_{p}}
        \mathrm{sim}\big(h_{\theta_s}(\hat G_i),c_t\big)
        \Big].
    \end{aligned}
\end{equation}
\endgroup
Let $\mathcal L_{\mathrm{low}}(\theta_s;\theta_g)$ denote the lower-level surrogate objective in Eq.~\eqref{eq:overall_loss}. We approximate the bi-level min-max problem with block-coordinate updates. For the adversarial block, with $\theta_g$, $\theta_s$, and $\mathbf p$ fixed, we update $\sigma$ by projected gradient ascent:
\begin{equation}
\small
    \label{eq:update_perturbation}
    \sigma^{t+1}
    =
    \Pi_{\tau_\sigma}\!\big(
    \sigma^t
    + \eta_\sigma
    \nabla_\sigma L_{c_{\mathrm{sem}}}(\theta_g^t,\sigma^t)
    \big),
\end{equation}
where $\Pi_{\tau_\sigma}$ is the Euclidean projection onto the $\ell_2$ ball $\{\sigma:\|\sigma\|_2\le\tau_\sigma\}$.
We then update the surrogate encoder on the lower-level objective:
\begin{equation}
\small
    \label{eq:update_surrogate_model}
    \theta_s^{t+1}
    =
    \theta_s^t
    -
    \eta_s
    \nabla_{\theta_s}
    \mathcal L_{\mathrm{low}}(\theta_s^t;\theta_g^t).
\end{equation}

With the updated surrogate fixed, we update the trigger generator by descending on the regularized upper-level graph objective:
\begin{equation}
\small
    \label{eq:update_graph_triggers}
    \theta_g^{t+1}
    =
    \theta_g^t
    -
    \eta_g
    \nabla_{\theta_g}
    \mathcal L_{\mathrm g}(\theta_g^t,\sigma^{t+1};\theta_s^{t+1}).
\end{equation}

Finally, with the graph-side variables fixed, we update the soft prompt:
\begin{equation}
\small
    \label{eq:update_soft_prompt}
    \mathbf p^{t+1}
    =
    \mathbf p^t
    -
    \eta_p
    \nabla_{\mathbf p}
    L_{\mathrm{atk}_t}(\mathbf p^t;\theta_g^{t+1},\theta_s^{t+1}),
\end{equation}
where $\eta_\sigma$, $\eta_s$, $\eta_g$, and $\eta_p$ are learning rates.
% The full training procedure is summarized in Alg.~\ref{alg:framework}. We initialize the surrogate graph encoder, trigger generator, perturbation vector, and soft prompt, then pre-train the surrogate with the graph-language alignment objective in Eq.~\eqref{eq:prelim_align}. Training first solves the graph-side bi-level min-max problem with adversarial, surrogate, and trigger-generator updates, then optimizes the text-side soft prompt with the graph side fixed. After convergence, the attacker obtains the optimized trigger generator $f_g$, readable trigger-node text $\mathcal T_i^g$, and soft prompt $\mathbf p$. During victim training, the attacker attaches the generated trigger subgraph to selected nodes, assigns $\mathcal T_i^g$ to trigger nodes, and inserts the fixed phrase $r$ into the associated node text; the victim then follows its standard GFM training pipeline.
The full training procedure is summarized in Alg.~\ref{alg:framework}. \method{} first pre-trains the surrogate graph encoder, solves the graph-side bi-level min-max problem through adversarial, surrogate, and trigger-generator updates, and then optimizes the text-side soft prompt with the graph side fixed. During victim training, the attacker attaches the generated trigger subgraph to poisoned nodes, assigns the readable trigger-node text $\mathcal T_i^g$, and inserts the fixed phrase $r$ into the node text; the victim follows its standard GFM training pipeline.

\begin{algorithm}[t!]
\caption{Training procedure of \method{}.}
\label{alg:framework}
\begin{algorithmic}[1]
\REQUIRE Clean TAG $\mathcal{G}$, labeled nodes $\mathcal V_l$, poisoned nodes $\mathcal V_p$, frozen text encoder $e$, target class $y_t$, text trigger $r$, trigger size $s$, candidate number $N_q$, hyperparameters $\lambda_{c_1},\lambda_{c_2},\tau_\sigma,M,N$.
\ENSURE Trigger generator $f_g$, trigger-node texts $\{\mathcal T_i^g\}$, and soft prompt $\mathbf p$.
\STATE Initialize $\theta_s,\theta_g,\sigma,\mathbf p$; pre-train $h_{\theta_s}$ by Eq.~\eqref{eq:prelim_align}.
\FOR{$t=1,\dots,N$}
    \STATE Generate $g_i=f_g(v_i)$ and $\hat G_i=a_g(v_i,g_i)$ for $v_i\in\mathcal V_p$.
\FOR{$m=1,\dots,M$}
    \STATE Update $\sigma$ by ascent on $\nabla_{\sigma} L_{c_{\mathrm{sem}}}$ using Eq.~\eqref{eq:update_perturbation}.
\ENDFOR
\STATE Update $\theta_s$ by descent on $\nabla_{\theta_s}\mathcal L_{\mathrm{low}}$ using Eq.~\eqref{eq:update_surrogate_model}.
\STATE Update $\theta_g$ by descent on $\nabla_{\theta_g}\mathcal L_{\mathrm g}$ using Eq.~\eqref{eq:update_graph_triggers}.
\ENDFOR
\STATE Build $\mathcal Q_{y_t}(T_i)$ by Eq.~\eqref{eq:text_candidate_pool} and retrieve $\mathcal T_i^g$ for $v_i\in\mathcal V_p$.
% \STATE Optimize $\mathbf p$ by Eq.~\eqref{eq:update_soft_prompt} until convergence.
\WHILE{the text-side objective has not converged}
    \STATE Update $\mathbf p$ by descent on $\nabla_{\mathbf{p}}L_{\mathrm{atk}_t}$ using Eq.~\eqref{eq:update_soft_prompt}.
\ENDWHILE
\RETURN $f_g$, $\{\mathcal T_i^g\}_{v_i\in\mathcal V_p}$, $\mathbf p$.
\end{algorithmic}
\end{algorithm}
\section{Experiments}
In this section, we evaluate the proposed \method{} on various TAG datasets to answer the following research questions: (i) \textbf{RQ1:} How does \method{} perform in backdoor attacks against various GFMs and defense settings? 
(ii) \textbf{RQ2:} Why does \method{} require both cross-modal coordination and trigger stealthiness?
(iii) \textbf{RQ3:} How do different components contribute to \method{}'s effectiveness?

\begin{table}[t]
    \centering 
    \setlength{\defaultaddspace}{0.1em} 
    \renewcommand{\arraystretch}{0.8} 
    \caption{\textbf{Dataset Statistics.}}
    \vspace{-0.5em}
    \label{tab:dataset_stats}
    \begin{tabular}{ccccc}
\toprule
\textbf{Dataset} & \textbf{Nodes} & \textbf{Edges} & \textbf{Text attribute} & \textbf{Classes} \\
\midrule
Cora  & 2,708 & 5,278 & Paper text / keywords & 7 \\
CiteSeer  & 3,312 & 4,552 & Title / abstract & 6 \\
WikiCS  & 11,701 & 216,123 & Article text & 10 \\
OGB-arxiv  & 169,343 & 1,166,243 & Title / abstract & 40 \\
\bottomrule
\end{tabular}
\end{table}

\begin{table*}[t]
\centering
\small
\caption{\textbf{Backdoor attack results across three victim GFMs and four TAG datasets without defenses (ASR (\%) $|$ ACC (\%)).} Only clean accuracy is reported for clean graphs.}
\vspace{-0.5em}
\label{tab:main_gfm_results}
\setlength{\tabcolsep}{4.0pt}
\renewcommand{\arraystretch}{0.9}
\begin{tabular}{llccccc}
\toprule
\textbf{Dataset} & \textbf{GFM} & \textbf{Clean Graph} & \textbf{CrossBA} & \textbf{PoisonPrompt} & \textbf{BadCLIP} & \textbf{Ours} \\
\midrule
\multirow{3}{*}{Cora}
& GraphCLIP & 67.30 & 32.20 $|$ 13.80 & 13.40 $|$ 49.30 & 7.30 $|$ 9.30 & \textbf{93.20 $|$ 68.30} \\
& GraphGPT  & 92.50 & 35.60 $|$ 73.20 & 12.30 $|$ 25.50 & 16.60 $|$ 46.40 & \textbf{97.70 $|$ 94.60} \\
& G2P2      & 62.50 & 33.20 $|$ 16.90 & 16.50 $|$ 54.60 & 12.60 $|$ 11.10 & \textbf{100.00 $|$ 59.60} \\
\midrule
\multirow{3}{*}{CiteSeer}
& GraphCLIP & 65.20 & 61.40 $|$ 68.50 & 10.30 $|$ 64.90 & 0.30 $|$ 16.80 & \textbf{92.40 $|$ 63.70} \\
& GraphGPT  & 83.10 & 29.60 $|$ 66.80 & 10.20 $|$ 16.90 & 19.60 $|$ 39.70 & \textbf{95.40 $|$ 83.60} \\
& G2P2      & 65.30 & 33.50 $|$ 56.90 & 11.40 $|$ 61.70 & 5.60 $|$ 23.80 & \textbf{100.00 $|$ 61.90} \\
\midrule
\multirow{3}{*}{WikiCS}
& GraphCLIP & 58.50 & 76.60 $|$ 59.50 & 4.50 $|$ 58.60 & 13.30 $|$ 14.70 & \textbf{100.00 $|$ 57.40} \\
& GraphGPT  & 67.70 & 34.30 $|$ 48.50 & 4.80 $|$ 14.30 & 18.30 $|$ 26.30 & \textbf{100.00 $|$ 67.60} \\
& G2P2      & 64.60 & 63.30 $|$ 54.60 & 9.30 $|$ 64.70 & 24.50 $|$ 34.90 & \textbf{100.00 $|$ 64.50} \\
\midrule
\multirow{3}{*}{OGB-arxiv}
& GraphCLIP & 31.20 & 37.30 $|$ 18.50 & 31.40 $|$ 21.60 & 3.30 $|$ 6.60 & \textbf{98.60 $|$ 28.90} \\
& GraphGPT  & 42.80 & 25.80 $|$ 11.30 & 1.70 $|$ 5.70 & 29.40 $|$ 17.10 & \textbf{93.60 $|$ 43.30} \\
& G2P2      & 22.30 & 42.70 $|$ 8.70 & 3.60 $|$ 20.70 & 10.10 $|$ 5.60 & \textbf{100.00 $|$ 18.90} \\
\bottomrule
\end{tabular}
\end{table*}

\subsection{Experimental Setups}
\label{sec:exp_setup}
\noindent\textbf{Datasets.}
We evaluate on four datasets widely used as text-attributed graphs in recent GFM studies~\cite{wang2024bridging,wen2023augmenting,zhu2025graphclip}: Cora, CiteSeer~\cite{sen2008collective}, WikiCS~\cite{mernyei2020wiki}, and OGB-arxiv~\cite{hu2020open}. Cora, CiteSeer, and OGB-arxiv are citation networks where nodes are papers and edges are citation links, with node texts derived from paper titles, abstracts, or document descriptions. WikiCS is a Wikipedia article graph where nodes are articles, edges are hyperlinks, and node texts are derived from article content. Dataset statistics are reported in Tab.~\ref{tab:dataset_stats}.

\noindent\textbf{Victim GFMs.}
We evaluate \method{} on three representative GFMs: GraphGPT~\cite{tang2024graphgpt}, GraphCLIP~\cite{zhu2025graphclip}, and G2P2~\cite{wen2023augmenting}. GraphGPT follows the LLM-as-predictor paradigm, whereas GraphCLIP and G2P2 follow the LLM-as-aligner paradigm. 
% Additional model details are in Appendix~\ref{sec:victim_model}.

\noindent\textbf{Baselines.}
We compare with three representative backdoor baselines: PoisonPrompt~\cite{yao2024poisonprompt}, BadCLIP~\cite{liang2024badclip}, and CrossBA~\cite{lyu2024crossba}. PoisonPrompt is a text-side backdoor attack, CrossBA is a graph-side backdoor attack, and BadCLIP is a multimodal backdoor method originally developed for CLIP, which we adapt to the TAG setting. 
% Details of the adaptations are provided in Appendix~\ref{appendix:baselines}.

\noindent\textbf{Defense Methods.}
We apply three backdoor defense strategies to help evaluate the stealthiness of backdoor attacks: Prune~\cite{dai2023unnoticeable}, Outlier Detection (OD)~\cite{bandyopadhyay2019outlier}, and DOMINANT~\cite{ding2019deep}. Prune and OD mainly target feature anomalies, whereas DOMINANT targets structural anomalies.

\noindent\textbf{Evaluation Protocol.}
We split each dataset into train, validation, and test sets with a ratio of 6:2:2, and keep the test set inaccessible during training. Unless otherwise stated, the poison rate is set to $0.4$ and the trigger size is set to $8$. For clean utility, we report clean accuracy (ACC) on the clean test set. For attack effectiveness, we report attack success rate (ASR), defined as the fraction of trigger-attached test nodes classified into the attacker-chosen target class $y_t$. All results are averaged over five runs.

\noindent\textbf{Implementation Details.}
Following existing GFM settings on TAGs~\cite{wen2023augmenting,zhu2025graphclip}, we encode node texts with frozen SBERT (all-MiniLM-L6-v2) into $384$-dimensional features. The victim graph encoders are graph transformers, with $12$ layers for GraphCLIP and $3$ layers for GraphGPT and G2P2. The attacker uses a $5$-layer GCN as the surrogate graph encoder, a $2$-layer MLP as the trigger generator, and a soft prompt of length $20$. Trigger-node texts are selected from $N_q=5$ target-class-styled candidates generated by Claude Haiku 4.5. For defenses, Prune removes edges with cosine similarity below $0.4$; OD removes the top $3\%$ nodes by autoencoder reconstruction loss; DOMINANT removes the top $3\%$ edges by adjacency reconstruction loss.

\begin{table*}[t]
\centering
\small
\caption{\textbf{Backdoor attack results under three backdoor defenses, averaged over three victim GFMs on Cora and OGB-arxiv (ASR (\%) $|$ ACC (\%)).} Only clean accuracy is reported for clean graphs.}
\vspace{-0.5em}
\label{tab:defense_avg_by_dataset}
\setlength{\tabcolsep}{4.0pt}
\renewcommand{\arraystretch}{0.9}
\begin{tabular}{llccccc}
\toprule
\textbf{Dataset} & \textbf{Defense} & \textbf{Clean Graph} & \textbf{CrossBA} & \textbf{PoisonPrompt} & \textbf{BadCLIP} & \textbf{Ours} \\
\midrule
\multirow{4}{*}{Cora}
& None     & \multirow{4}{*}{74.10} & 33.67 $|$ 34.63 & 14.07 $|$ 43.13 & 12.17 $|$ 22.27 & \textbf{96.97 $|$ 74.17} \\
& Prune    &     & 29.27 $|$ 33.53 & 10.30 $|$ 42.37 & 8.53 $|$ 21.33 & \textbf{95.20 $|$ 72.10} \\
& OD       &     & 29.70 $|$ 33.90 & 11.13 $|$ 42.70 & 7.63 $|$ 21.27 & \textbf{96.63 $|$ 73.50} \\
& DOMINANT &     & 23.77 $|$ 33.30 & 8.50 $|$ 43.10 & 7.33 $|$ 20.33 & \textbf{96.90 $|$ 73.57} \\
\midrule
% \multirow{4}{*}{CiteSeer}
% & None     & \multirow{4}{*}{71.20} & 41.50 $|$ 64.07 & 10.63 $|$ 47.83 & 8.50 $|$ 26.77 & \textbf{95.93 $|$ 69.73} \\
% & Prune    &     & 33.00 $|$ 63.07 & 10.27 $|$ 45.97 & 7.10 $|$ 23.60 & \textbf{92.80 $|$ 68.03} \\
% & OD       &     & 35.50 $|$ 62.60 & 11.53 $|$ 46.63 & 4.07 $|$ 24.93 & \textbf{93.37 $|$ 69.03} \\
% & DOMINANT &     & 12.83 $|$ 63.30 & 10.17 $|$ 46.77 & 2.60 $|$ 25.53 & \textbf{95.47 $|$ 69.43} \\
% \midrule
% \multirow{4}{*}{WikiCS}
% & None     & \multirow{4}{*}{63.60} & 58.07 $|$ 54.20 & 6.20 $|$ 45.87 & 18.70 $|$ 25.30 & \textbf{100.00 $|$ 63.17} \\
% & Prune    &     & 51.90 $|$ 54.30 & 5.07 $|$ 43.37 & 14.73 $|$ 23.17 & \textbf{100.00 $|$ 62.20} \\
% & OD       &     & 52.57 $|$ 54.33 & 6.07 $|$ 44.83 & 14.60 $|$ 23.93 & \textbf{100.00 $|$ 62.80} \\
% & DOMINANT &     & 24.80 $|$ 54.33 & 6.10 $|$ 44.23 & 8.60 $|$ 23.90 & \textbf{100.00 $|$ 62.93} \\
% \midrule
\multirow{4}{*}{OGB-arxiv}
& None     & \multirow{4}{*}{32.10} & 35.27 $|$ 12.83 & 12.23 $|$ 16.00 & 14.27 $|$ 9.77 & \textbf{97.40 $|$ 30.37} \\
& Prune    &     & 31.43 $|$ 12.70 & 17.03 $|$ 10.87 & 8.60 $|$ 9.57 & \textbf{96.60 $|$ 28.20} \\
& OD       &     & 26.50 $|$ 12.73 & 16.77 $|$ 11.23 & 9.20 $|$ 9.50 & \textbf{97.33 $|$ 29.70} \\
& DOMINANT &     & 12.07 $|$ 12.77 & 16.97 $|$ 11.20 & 4.67 $|$ 9.53 & \textbf{97.30 $|$ 29.97} \\
\bottomrule
\end{tabular}
\end{table*}

\subsection{Attack Results}
\label{sec:backdoor_performance}

To answer \textbf{RQ1}, we compare \method{} with baselines on three representative GFMs and four TAG datasets under various defense settings in terms of attack performance and stealthiness.

\noindent\textbf{Comparisons with Baseline Backdoor Attacks.}
We conduct experiments on four real-world TAGs against three victim GFMs (i.e., GraphCLIP, GraphGPT and G2P2). Other settings follow Sec.~\ref{sec:exp_setup}. Tab.~\ref{tab:main_gfm_results} reports the results without defenses. From the table, we make the following observations:
\begin{itemize}[leftmargin=*]
    \item \emph{Existing backdoor attacks perform poorly on GFMs}. Averaged over all datasets and victim GFMs, CrossBA, PoisonPrompt, and BadCLIP achieve only $42.12\%$, $10.78\%$, and $13.41\%$ ASR, respectively, often with ACC degradation, supporting our claim in Sec.~\ref{subsec:limitation_existing_backdoor_GFMs} that single-side adaptations are insufficient for GFMs on TAGs.
    \item \emph{\method{} achieves the best ASR across all datasets and GFMs while maintaining competitive ACC.} It reaches at least $92.40\%$ ASR in every setting with limited ACC degradation, validating the effectiveness of \method{} against GFMs.
    \item \emph{The largest gaps arise when baselines perturb only one modality}. For example, on Cora with GraphCLIP, PoisonPrompt preserves moderate ACC but reaches only $13.40\%$ ASR, while CrossBA improves ASR to $32.20\%$ but drops ACC to $13.80\%$; \method{} achieves $93.20\%$ ASR with $68.30\%$ ACC, supporting the need for cross-modal coordination. We further analyze this mechanism in Sec.~\ref{subsec:depth_analysis_method}.
\end{itemize}

\noindent\textbf{Comparisons under Backdoor Defenses.}
To evaluate the stealthiness of \method{}, we compare it with baselines under three defense methods: Prune, OD, and DOMINANT. Other settings follow Sec.~\ref{sec:exp_setup}. Tab.~\ref{tab:defense_avg_by_dataset} reports average results over the three victim GFMs on Cora and OGB-arxiv datasets. From the table, \method{} outperforms all baselines under all defense strategies. Specifically, \method{} keeps at least $95.20\%$ ASR, whereas the baselines degrade substantially, especially under DOMINANT. These defense results indicate the stealthiness of \method{}: readable trigger-node text reduces feature-level anomaly signals, while structural regularization keeps the trigger-attached subgraph close to the original local structure. We further analyze trigger stealthiness in depth in Sec.~\ref{subsec:depth_analysis_method}.

\subsection{In-depth Analysis of \method{}}
\label{subsec:depth_analysis_method}
To answer \textbf{RQ2}, we conduct in-depth analysis of two properties behind \method{} beyond ASR and ACC: \emph{cross-modal coordination}, which drives poisoned embeddings into the target-class region, and \emph{trigger stealthiness}, which keeps the trigger inconspicuous in both text and structure.

\begin{figure*}[t]
  \centering
  \includegraphics[width=0.8\textwidth]{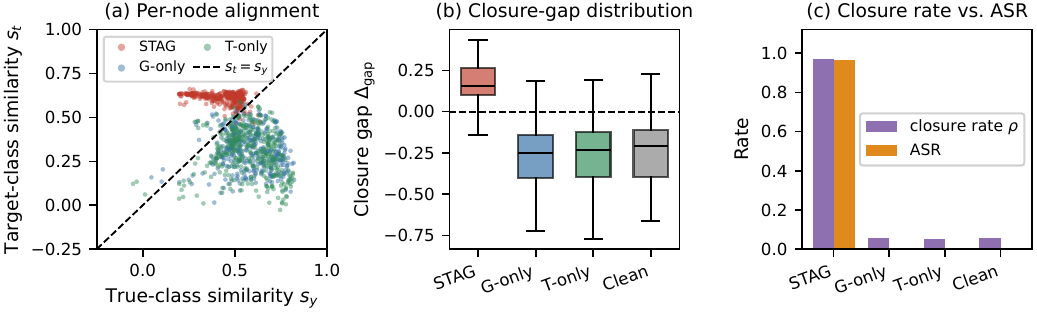}
  \vspace{-0.5em}
\caption{\textbf{Embedding-closure analysis on Cora with GraphCLIP.} (a) Similarity to the target-class anchor ($s_t$) vs. the true-class anchor ($s_y$). (b) Closure-gap distribution. (c) Closure rate $\rho$ vs. ASR.}
  \label{fig:closure}
\end{figure*}

\noindent\textbf{Coordination Analysis.}
We analyze why single-modal backdoors fail and why cross-modal coordination works by tracing poisoned graph embeddings under graph-language alignment. We compare four diagnostic settings: clean baseline, graph-only attack, text-only attack, and \method{}. Here, graph-only uses only the trigger subgraph with clean text, while text-only uses only the text trigger without a graph trigger. For each attacked node, let $z_i$ be its graph embedding, and let $e(y_t)$ and $e(y_i)$ be the target-class and true-class text anchors. We measure cosine similarities $s_t=\cos(z_i,e(y_t))$ and $s_y=\cos(z_i,e(y_i))$, and define the closure gap as $\Delta_{\mathrm{gap}}=s_t-s_y$.  $\Delta_{\mathrm{gap}} >0$ means that $z_i$ is closer to the target-class anchor; the closure rate $\rho=\Pr[\Delta_{\mathrm{gap}}>0]$ tracks whether poisoned embeddings enter the target-class text region. Fig.~\ref{fig:closure} reports the results on Cora with GraphCLIP. We observe:
\begin{itemize}[leftmargin=*]
    \item \emph{Single-modal attacks fail to enter the target-class text region.} Their embeddings mostly stay below the diagonal (Fig.~\ref{fig:closure}(a)) and have negative closure gaps (Fig.~\ref{fig:closure}(b)). The graph-only trigger has $\Delta_{\mathrm{gap}}=-0.27$ and $\rho=0.06$, while the text-only trigger is close to clean training ($\rho=0.05$ vs. $0.06$; Fig.~\ref{fig:closure}(c)). It indicates that graph-only triggers are pulled back by clean-text alignment, whereas text-only triggers do not move the graph embedding being scored.
    \item \emph{\method{} moves poisoned embeddings into the target-class text region.} Its embeddings shift above the diagonal and obtain positive closure gaps ($\Delta_{\mathrm{gap}}=0.18$), yielding a high closure rate close to ASR ($\rho=0.97$). This indicates that coordination aligns both sides: the text-side prompt removes the clean-text pull-back, while the graph trigger moves the scored representation toward the target class.
\end{itemize}

\noindent\textbf{Stealthiness Analysis.}
We evaluate trigger stealthiness from textual and structural perspectives. For text, we use PPL~\cite{radford2019language} to measure fluency under a frozen GPT-2, and ONION~\cite{qi2021onion} to measure the fraction of trigger words removed by a perplexity-based textual backdoor detector. For structure, we use $|\Delta\bar d|$~\cite{xi2021graph} to measure the mean-degree shift between the clean $K$-hop subgraph $G_i$ and the trigger-attached subgraph $\hat G_i$, and the Kolmogorov-Smirnov statistic~\cite{massey1951kolmogorov} to measure the maximum distance between their degree distributions. Lower values indicate stronger stealthiness. We compare \method{} with baselines and two variants: w/o $L_{c_{\mathrm{sem}}}$ and w/o $L_{c_{\mathrm{str}}}$ remove semantic and structural concealment, respectively. Tab.~\ref{tab:stealth} reports the results on Cora with GraphCLIP. We observe:
\begin{itemize}[leftmargin=*]
\item \emph{\method{} yields readable trigger-node text.} Its PPL is close to clean text ($58.4$ vs. $53.4$), far below text baselines, and ONION removes no trigger words. Without $L_{c_{\mathrm{sem}}}$, PPL rises to $20712.7$ and ONION removal to $45.2\%$, showing that semantic concealment enables natural text realization.
    \item \emph{\method{} preserves structural plausibility.} It keeps degree shift and KS small ($|\Delta\bar d|=0.66$, KS $=0.27$). Removing $L_{c_{\mathrm{str}}}$ increases them to $4.39$ and $0.94$, matching CrossBA's deviation. This shows that structural concealment keeps the trigger-attached subgraph close to the clean local structure.
\end{itemize}

\begin{table}[t]
\centering
\small
\caption{\textbf{Trigger-stealthiness measurements on Cora with GraphCLIP.} 
% Clean text is the benign node-text reference. 
Lower values indicate stronger stealthiness.}
\vspace{-0.5em}
\label{tab:stealth}
\setlength{\tabcolsep}{3.2pt}
\renewcommand{\arraystretch}{0.92}
\begin{tabular}{lcccc}
\toprule
& \multicolumn{2}{c}{\textbf{Textual}} & \multicolumn{2}{c}{\textbf{Structural}} \\
\cmidrule(lr){2-3}\cmidrule(lr){4-5}
\textbf{Method} & PPL & ONION\% & $|\Delta\bar d|$ & KS \\
\midrule
Clean text & 53.4 & 0.0 & -- & -- \\
\midrule
CrossBA & -- & -- & 4.39 & 0.94 \\
PoisonPrompt & 67083.8 & 33.1 & -- & -- \\
BadCLIP & 441.5 & 0.8 & -- & -- \\
\midrule
\method{} w/o $L_{c_{\mathrm{sem}}}$ & 20712.7 & 45.2 & 0.66 & 0.27 \\
\method{} w/o $L_{c_{\mathrm{str}}}$ & 58.4 & 0.0 & 4.39 & 0.94 \\
\method{} & \textbf{58.4} & \textbf{0.0} & \textbf{0.66} & \textbf{0.27} \\
\bottomrule
\end{tabular}
\end{table}

\subsection{Ablation Studies}
\begin{figure}[t]
    \centering
    \begin{subfigure}[t]{0.48\linewidth}
        \centering
        \includegraphics[width=0.9\linewidth]{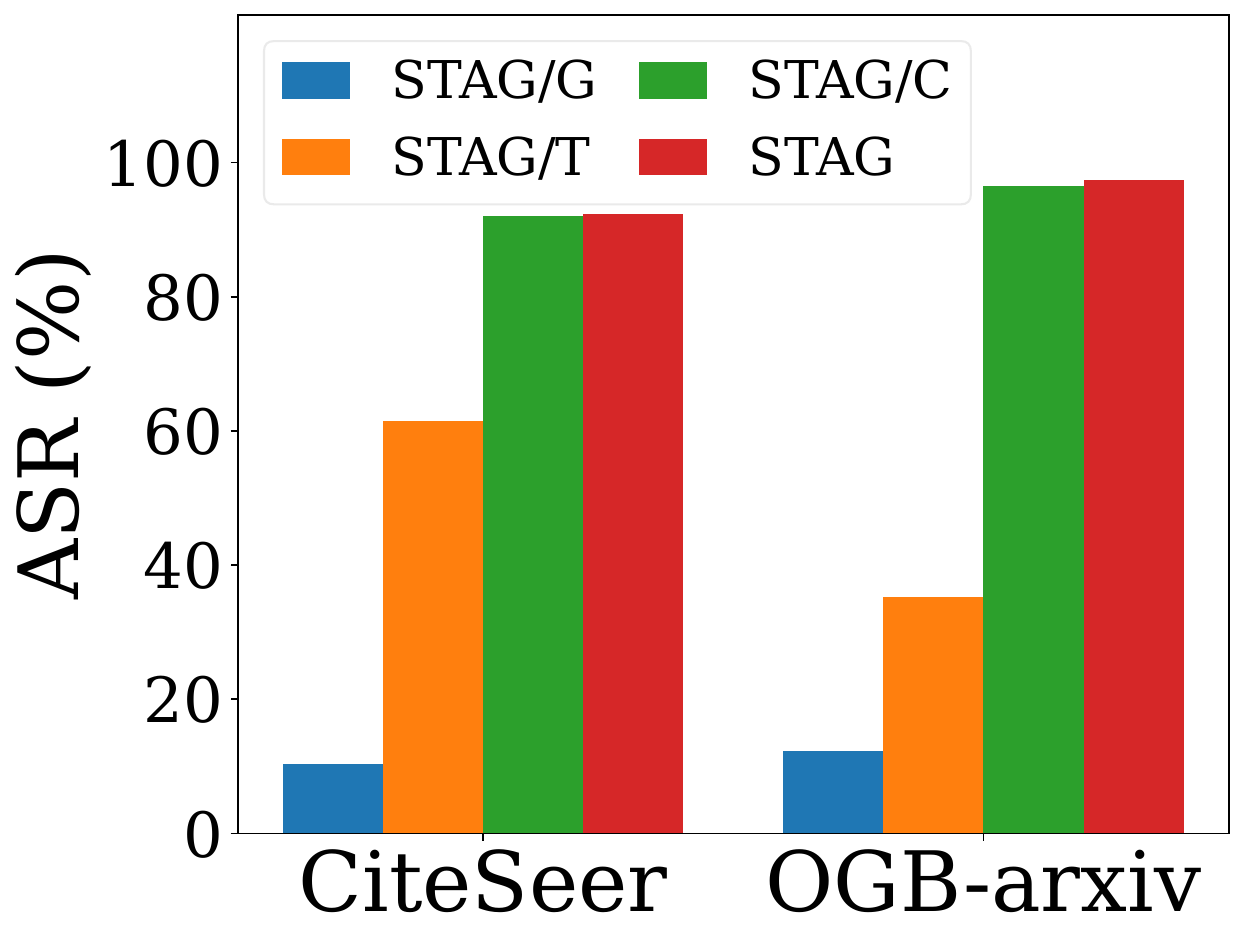}
            \vspace{-0.5em}
        \caption{No defense}
        \label{fig:ablation_no_defense}
    \end{subfigure}
    \vspace{-0.5em}
    % \hfill
    \begin{subfigure}[t]{0.48\linewidth}
        \centering
        \includegraphics[width=0.9\linewidth]{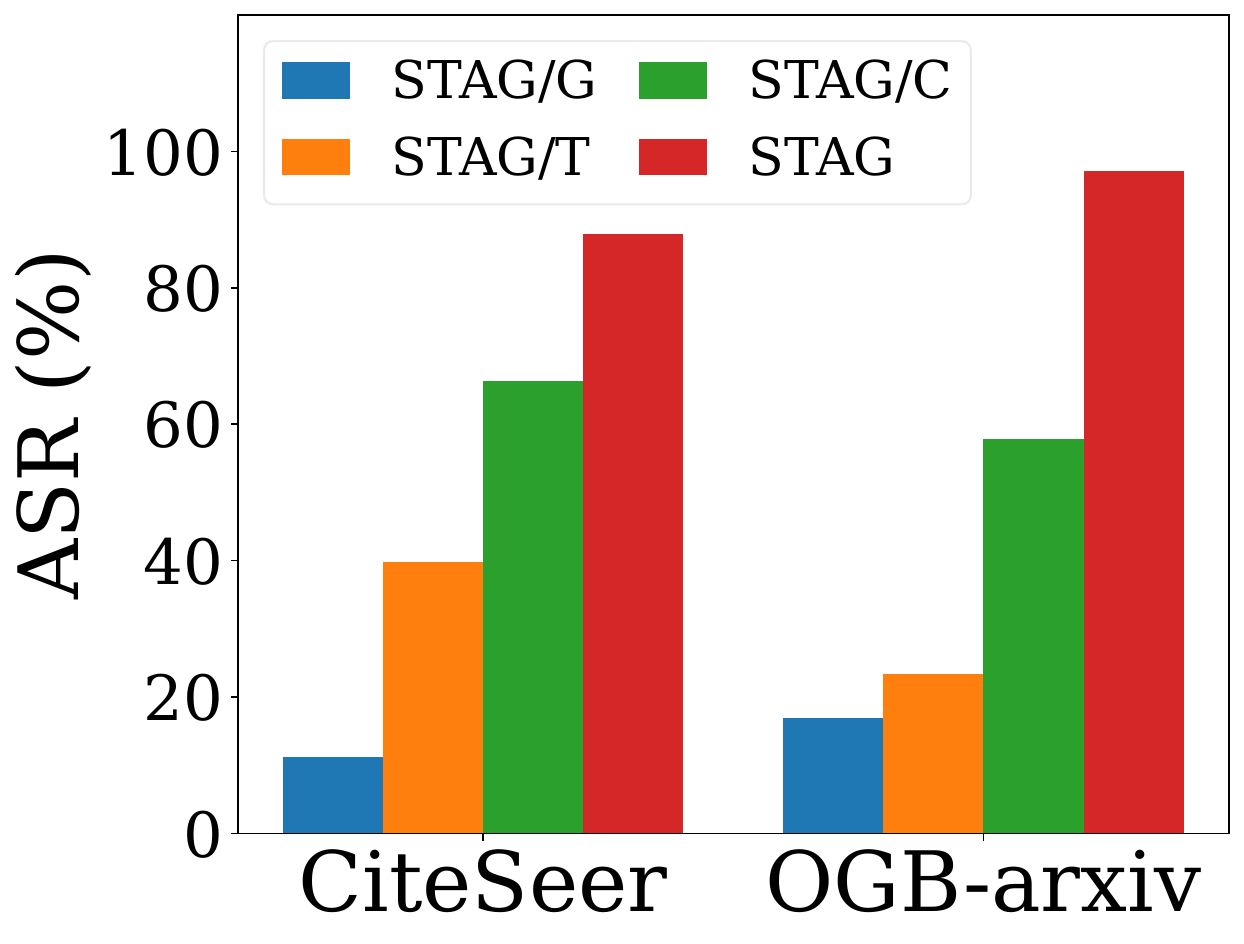}
            \vspace{-0.5em}
        \caption{Avg. defense}
        \label{fig:ablation_avg_defense}
    \end{subfigure}
    \caption{\textbf{Ablation studies on CiteSeer and OGB-arxiv.}}
    \label{fig:ablation_study}
\end{figure}

To answer \textbf{RQ3}, we conduct ablation studies to understand the effects of the attack-side and concealment components of \method{}. We compare the full \method{} with three ablated variants: \textbf{STAG/G} removes the graph-side attack objective $L_{\mathrm{atk}_g}$; \textbf{STAG/T} removes the text-side attack objective $L_{\mathrm{atk}_t}$; and \textbf{STAG/C} averages two concealment ablations that remove $L_{c_{\mathrm{sem}}}$ or $L_{c_{\mathrm{str}}}$. Fig.~\ref{fig:ablation_study} reports ASR on CiteSeer and OGB-arxiv under no defense and averaged over three defenses, namely Prune, OD, and DOMINANT. We observe:
\begin{itemize}[leftmargin=*]
    \item \emph{\method{} performs best across settings.}
    \method{} achieves the highest ASR under both no-defense and defense settings. Removing either $L_{\mathrm{atk}_g}$ or $L_{\mathrm{atk}_t}$ substantially reduces ASR, showing that single-side backdoor signals are insufficient for attacking graph-language alignment in GFMs.
    \item \emph{The graph-side objective is especially critical.}
    Removing $L_{\mathrm{atk}_g}$ causes the largest drop. This is consistent with our analysis in Sec.~\ref{subsec:depth_analysis_method}: without the graph-side objective, the text-side trigger cannot by itself move graph embeddings into the target-class text region.
    \item \emph{Concealment improves robustness without weakening the attack.}
    STAG/C achieves ASR close to \method{} without defenses, but its ASR drops under defenses, especially on OGB-arxiv. This indicates that the concealment losses do not reduce attack effectiveness while improving robustness against defenses targeting feature and structural anomalies.
\end{itemize}

\subsection{Impact of the Trigger Size}
\begin{figure}[t]
    \centering
        \begin{subfigure}[t]{0.48\linewidth}
        \centering
        \includegraphics[width=0.9\linewidth]{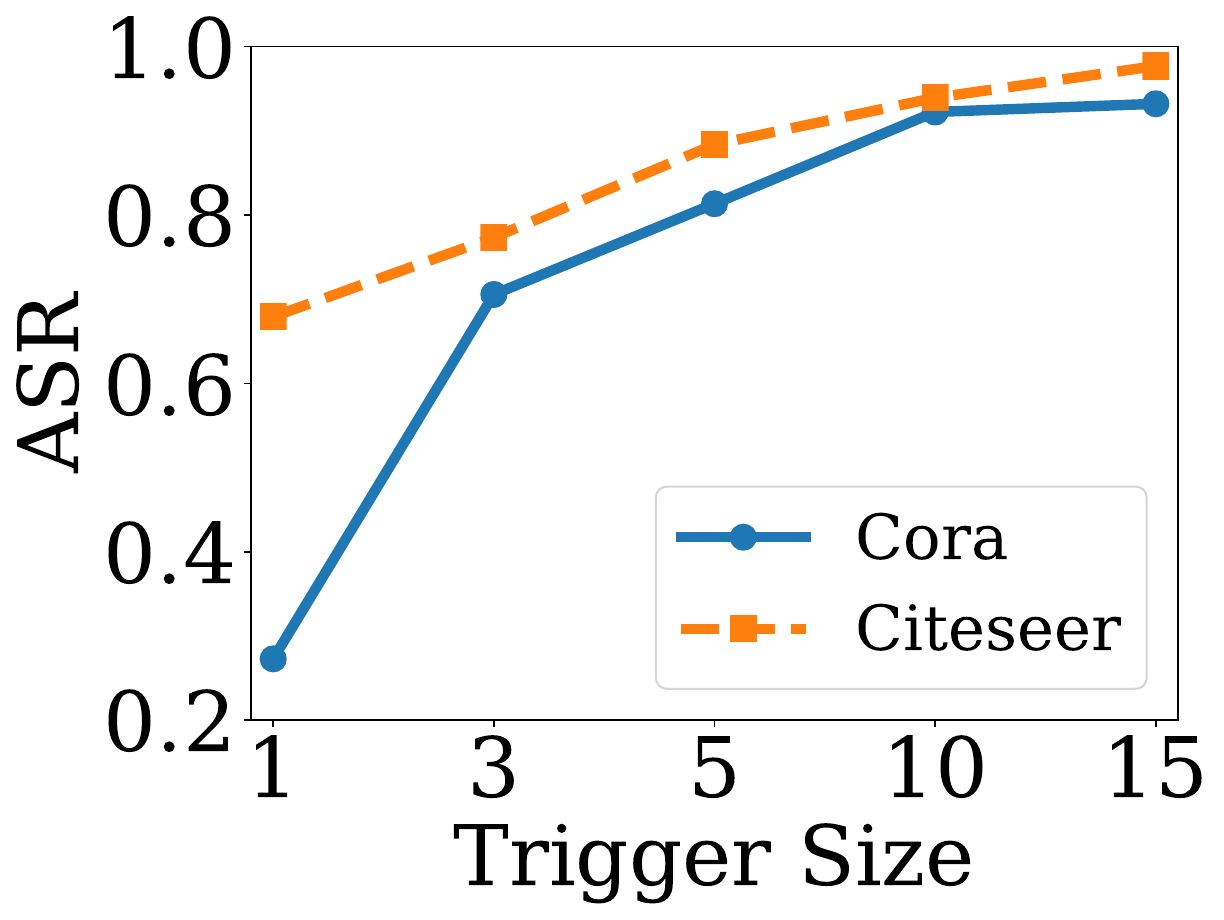}
        \vspace{-0.5em}
        \caption{ASR}
        \label{fig:trigger_size_asr}
    \end{subfigure}
    \hfill
    \begin{subfigure}[t]{0.48\linewidth}
        \centering
        \includegraphics[width=0.9\linewidth]{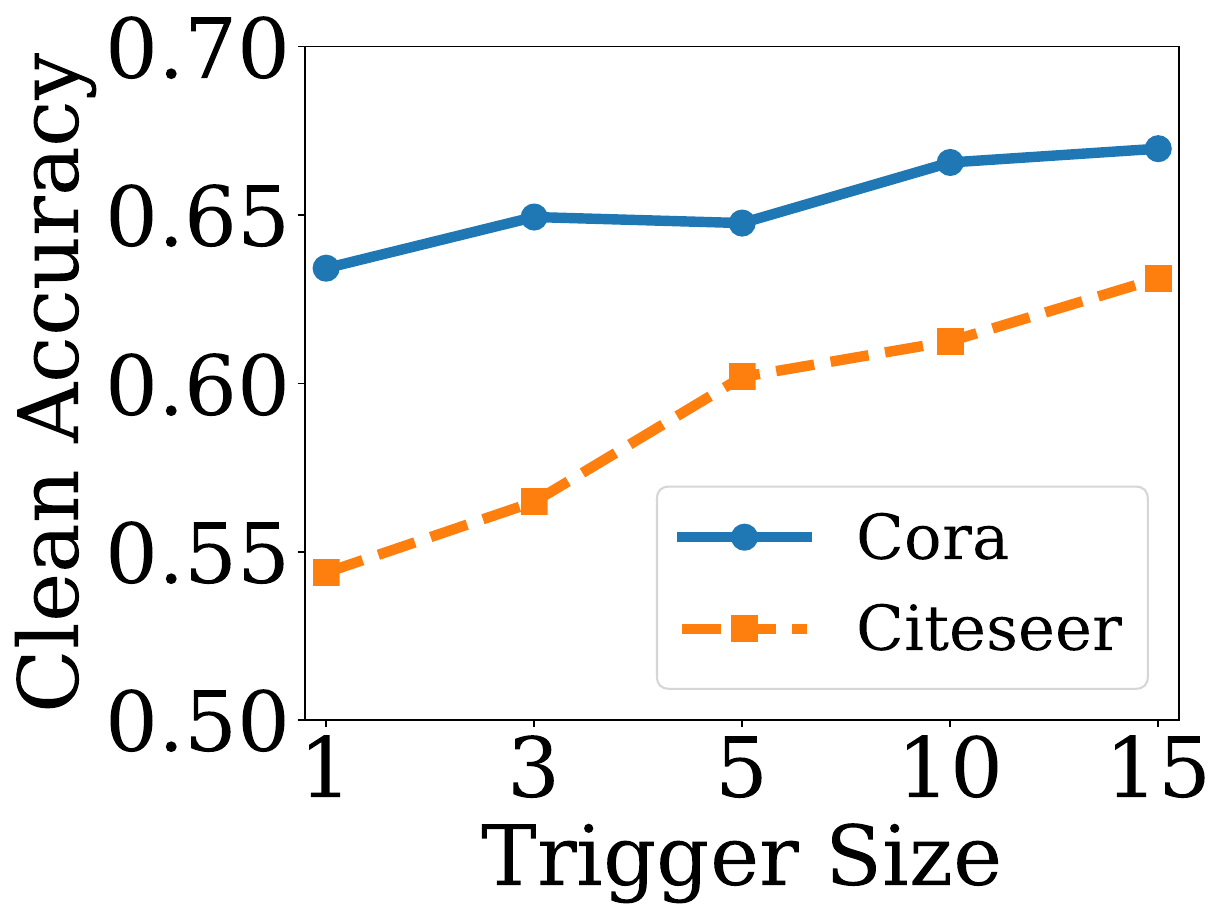}
        \vspace{-0.5em}
        \caption{ACC}
        \label{fig:trigger_size_acc}
    \end{subfigure}
    \vspace{-0.5em}
    \caption{\textbf{Impact of the Size of Trigger Nodes.}}
    \label{fig:trigger_size_overall}
\end{figure}

We study how the graph-trigger-size budget affects \method{}'s attack performance. Specifically, we vary the number of trigger nodes in $\{1,3,5,10,15\}$. The other settings are the same as in Sec.~\ref{sec:exp_setup}. Fig.~\ref{fig:trigger_size_overall} reports ACC and ASR on Cora and CiteSeer.
As the trigger size increases, ASR increases on both datasets, indicating that a larger trigger graph provides a stronger backdoor signal. ACC remains stable, with fluctuations within $5\%$. This indicates that \method{} can improve attack effectiveness with a limited trigger-size budget while preserving clean utility.

\subsection{Impact of the Poison Rate}
\begin{figure}[t]
    \centering
    \begin{subfigure}[t]{0.48\linewidth}
        \centering
        \includegraphics[width=0.95\linewidth]{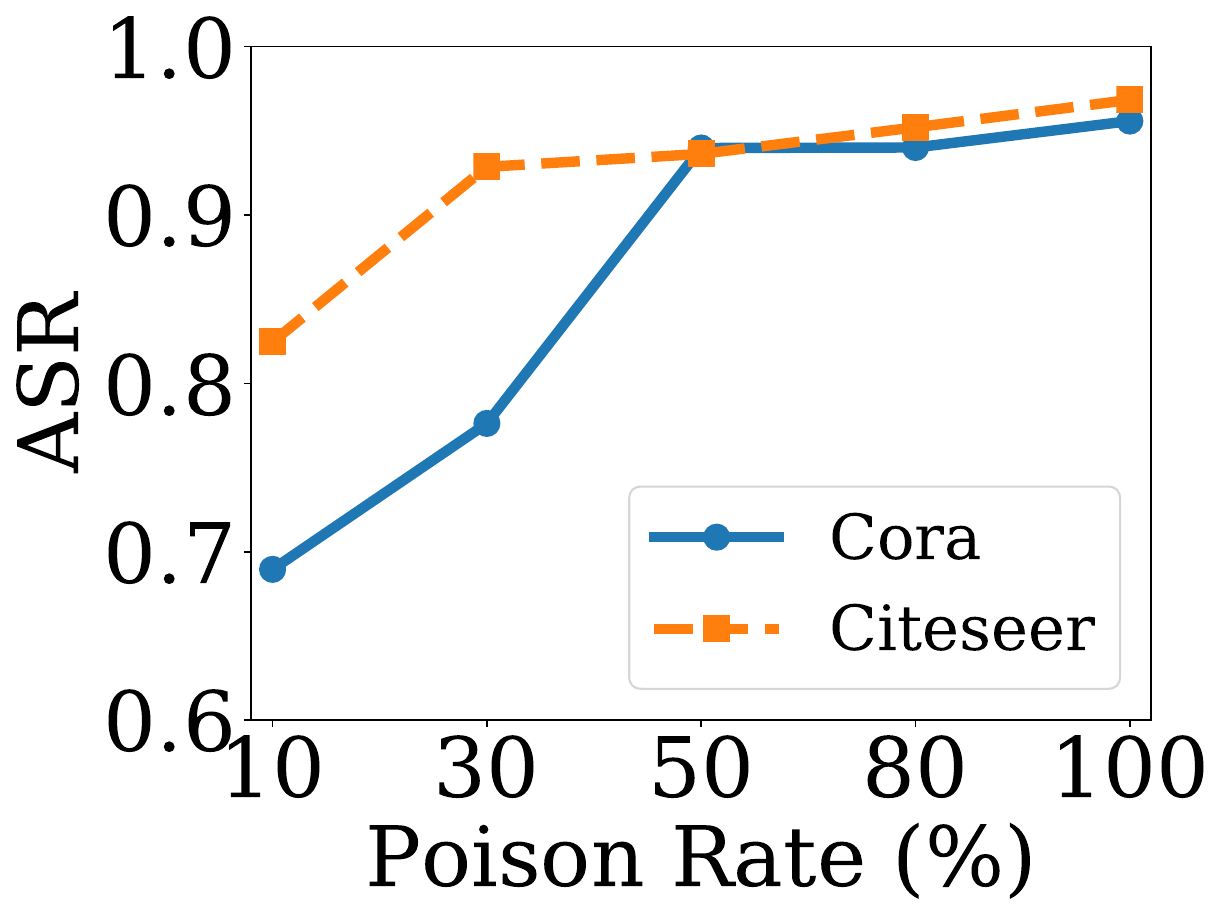}
        \vspace{-0.5em}
        \caption{ASR}
        \label{fig:poison_rate_asr}
    \end{subfigure}
    \hfill
    \begin{subfigure}[t]{0.48\linewidth}
        \centering
        \includegraphics[width=0.95\linewidth]{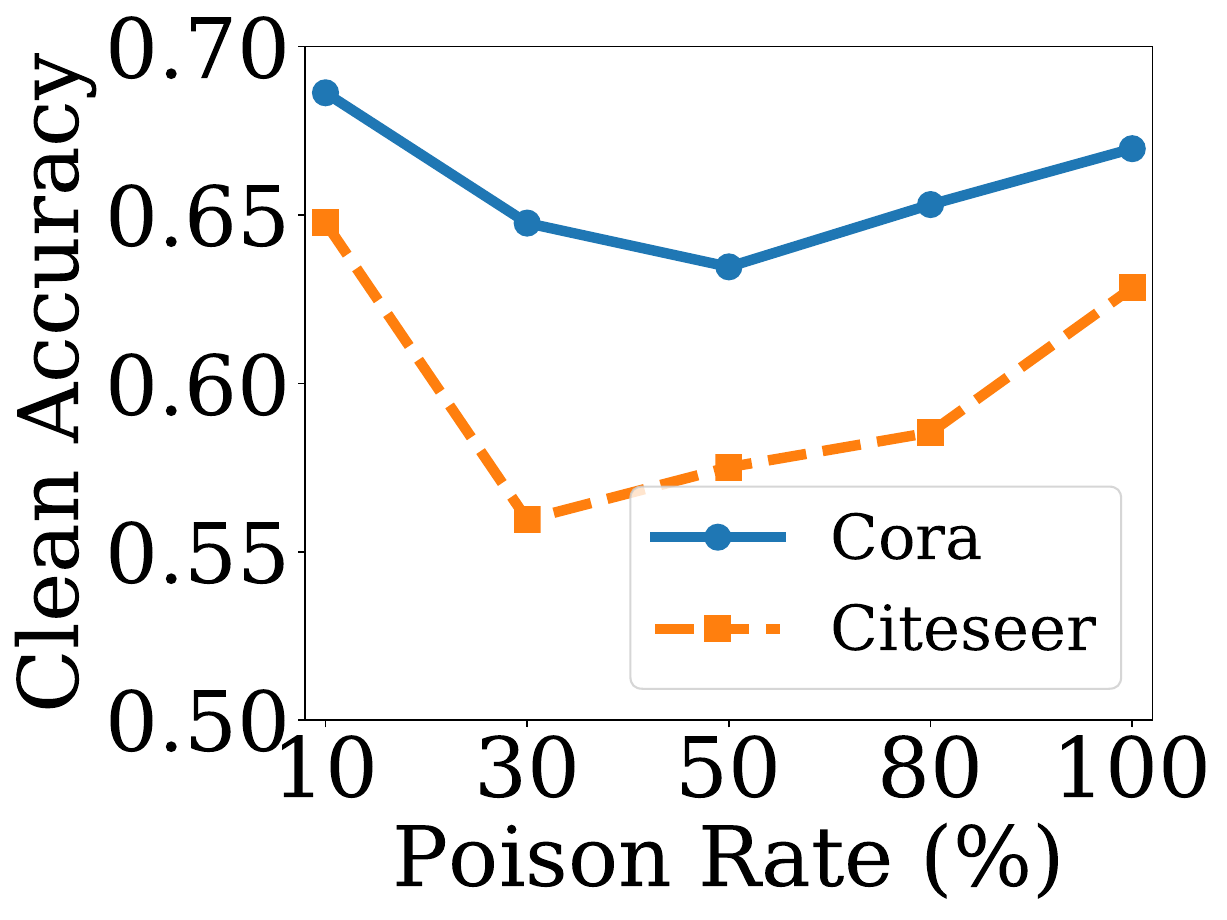}
        \vspace{-0.5em}
        \caption{ACC}
        \label{fig:poison_rate_acc}
    \end{subfigure}
    \vspace{-0.5em}
    \caption{\textbf{Impact of the Poison Rate.}}
    \label{fig:poison_rate_overall}
\end{figure}

We also conduct experiments to assess the attack performance of \method{} under different budgets for the ratio of poisoned nodes. Specifically, we vary the poison rate in $\{10\%,30\%,50\%,80\%,100\%\}$. Other settings follow Sec.~\ref{sec:exp_setup}. Fig.~\ref{fig:poison_rate_overall} reports ASR and ACC on Cora and CiteSeer. We observe that ASR is already high under small poison rates and further increases as the poison rate grows, while ACC fluctuates within a bounded range. This indicates the effectiveness of \method{} even under limited attack budgets.

\section{Conclusion}
In this paper, we investigate a novel problem of backdoor attacks against the graph-language alignment interface of GFMs on TAGs. We propose \method{}, which coordinates a graph-trigger generator with a text-side soft prompt so that trigger-attached subgraphs and triggered texts move toward the same target-class text region during graph-language alignment. To make the attack viable on TAGs, \method{} realizes trigger-node attributes as readable text and regularizes trigger-attached subgraphs to preserve local structural plausibility.
Extensive experiments across TAG datasets, victim GFMs, and defenses show that \method{} achieves high attack success while maintaining clean accuracy.
Two directions remain open: extending the analysis to broader graph-LLM pipelines, such as graph retrieval-augmented generation, and developing effective defenses against alignment-level backdoor attacks on GFMs.

\section{Acknowledgment}
This material is based upon work supported by, or in part by NSF award \#2555559. 
The views and conclusions contained in this material are those of the authors and should not be interpreted as necessarily representing the official policies, either expressed or implied, of the funding agencies.

\bibliographystyle{IEEEtran}
\bibliography{main}

% \newpage
% \appendix
% \input{8_Appendix_v2}

% \vspace{12pt}
% \color{red}
% IEEE conference templates contain guidance text for composing and formatting conference papers. Please ensure that all template text is removed from your conference paper prior to submission to the conference. Failure to remove the template text from your paper may result in your paper not being published.

\end{document}